\documentclass[12pt]{article}
\usepackage{graphicx}
\usepackage{amsmath}
\usepackage{amsfonts}
\usepackage{geometry}

\usepackage{natbib}

\ifx\pdfoutput\@undefined\usepackage[usenames,dvips]{color}
\else\usepackage[usenames,dvipsnames]{color}
\IfFileExists{pdfcolmk.sty}{\usepackage{pdfcolmk}}{} 
\fi
\usepackage[plainpages=false,pdfpagelabels,pagebackref=false,naturalnames=true,hyperindex=true,pdftitle={Computing Networks},pdfauthor={Carlos Gershenson}]{hyperref}
\hypersetup{colorlinks=true,
urlcolor=Cerulean,linkcolor=BrickRed,citecolor=RoyalBlue,a4paper,
  pdfpagemode=None,
  pdfstartview=FitH}
\usepackage[all]{hypcap}

\begin{document}

\title{Computing Networks:\\ A General Framework to Contrast\\ Neural and Swarm Cognitions}

\author{Carlos Gershenson \\
Computer Sciences Department \\
Instituto de Investigaciones en Matem\'aticas Aplicadas y en Sistemas \\
Universidad Nacional Aut\'onoma de M\'exico\\
Ciudad Universitaria, A.P. 20-726\\
01000 M\'exico D.F. M\'exico\\
\href{mailto:cgg@unam.mx}{cgg@unam.mx} \
\url{http://turing.iimas.unam.mx/~cgg}
}

\maketitle

\begin{abstract}
This paper presents the Computing Networks (CNs) framework. CNs are used to generalize neural and swarm architectures. Artificial neural networks, ant colony optimization, particle swarm optimization, and realistic biological models are used as examples of instantiations of CNs. The description of these architectures as CNs allows their comparison. Their differences and similarities allow the identification of properties that enable neural and swarm architectures to perform complex computations and exhibit complex cognitive abilities. In this context, the most relevant characteristics of CNs are the existence multiple dynamical and functional scales. The relationship between multiple dynamical and functional scales with adaptation, cognition (of brains and swarms) and computation is discussed.


\end{abstract}

\section{Introduction}
\label{intro}

The complex behavior exhibited by swarms has been actively studied in recent decades \citep{hoelldobler1990ants,Aron:1990,Reznikova:2007,Ryabko:2009} and exploited in engineering \citep{BonabeauEtAl1999,DorigoStuetzle2004,DorigoEtAl2004}. Recent research has highlighted the similarities between swarms and brains, noting that swarms are capable of performing cognitive tasks \citep{Chialvo:1995,Couzin:2009,Marshall:2009,Passino:2008,Trianni-Tuci:09:ecal}. Contributing to the effort of understanding these similarities, with biological and engineering aims, this paper generalizes models of swarm and neural architectures. In particular, artificial neural networks (ANNs), ant colony optimization (ACO), and particle swarm optimization (PSO) are described under the same general framework. The generalization, named \emph{computing networks} (CNs), provides a common ground for comparison and for studying the underlying mechanisms and the  computational properties common to neural and swarm architectures. As a guiding principle, we can say that neural and swarm architectures compute ``unknown" functions $f$, i.e. they explore phase spaces of functions until a satisfactory $f$ is found according to certain criteria.

Swarm cognition \citep{Trianni-Tuci:09:ecal} studies the intersection of the scientific study of natural swarms and neural cognition, with the aim of increasing our understanding of cognition relating it to the self-organization of swarms. CNs provide a general framework to contrast the cognition exhibited by brains and swarms. This particular aim for defining CNs restricts their usefulness, i.e.~the purpose of CNs is to increase our understanding of cognitive architectures, not to produce better models or more powerful computational algorithms.

In the next section, the computing networks are defined. In the following sections, CNs are used to describe ANNs, ACO \& PSO. These architectures were chosen for their generality and widespread use. More realistic biological models are also presented in terms of CNs within these sections.
This is followed by a comparison and discussion. In this section, similarities and differences of the architectures are explored, followed by the discussion multiple dynamical and functional scales. Also, the suitability and equivalence of different architectures is considered. The discussion continues dealing with the cognition of swarm and neural architectures, followed by an examination of alternate descriptions of the architectures. Conclusions close the paper.

\section{Computing Networks: A General Descriptive Framework}
\label{sec:1}

Many systems can be described as networks, i.e.~nodes connected by edges \citep{Newman2003,NewmanEtAl2006}. In this paper, we use the concept of \emph{computing network} (CN) as a generalization of artificial neural networks  \citep{RumelhartEtAl1986,Hopfield:1988}, ant colony optimization \citep{Dorigo:1991,DorigoStuetzle2004,Dorigo:2005,Dorigo:2007}, and particle swarm optimization \citep{Kennedy:1995,Kennedy:2001,Dorigo:2008}. In this way, the similarities and differences between these characteristic models of neural and swarm intelligence are studied under the same formalism.

A computing network $C(N,K,a,f)$ is defined as \textbf{a set of nodes $N$ linked by a set of edges $K$ used by an algorithm $a$ to compute a function $f$. Nodes and edges can have internal variables that determine their state, and functions that determine how their state changes}. This is a very general definition, and can be applied to describe many architectures and models beyond those discussed in this paper. Computing networks can be stochastic or deterministic (depending on the determinism of functions and algorithms), synchronous or asynchronous (depending on the updating used for the change of states of nodes and edges \citep{Gershenson2002e,Gershenson2004b}), discrete \citep{Wuensche1998} or continuous (depending on the type of variables of nodes and edges).

\section{Artificial Neural Networks}
\label{sec:2}

Artificial Neural Networks (ANNs) were originally proposed as logical models of the neocortex \citep{mcculloch:1943}. However, their computing power \citep{hopfield1982neural} has shifted the research focus from their plausibility as neural models to their application in different fields. There are many different types of ANNs, with different properties and implementations~\citep{RumelhartEtAl1986,Kohonen2000}. Here there will be no focus on any particular type of ANN.

\begin{figure}[htbp]
\begin{center}
\includegraphics[width=0.5\textwidth]{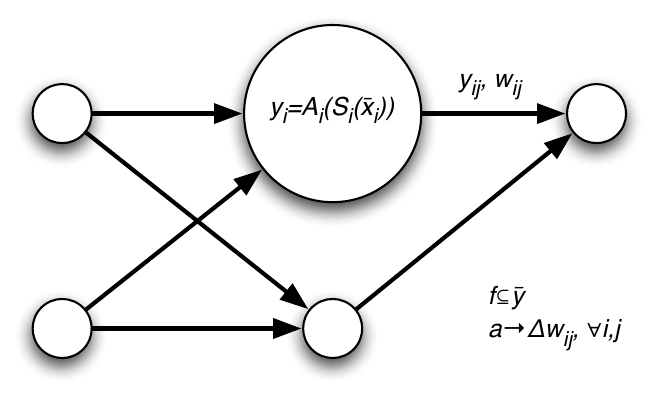}
\caption{Schematic of an ANN instantiation of a CN. Nodes have a function $y_i$ that is computed from its inputs ($\bar{x}_i$). Edges have weights $w_{ij}$ to determine the importance of the interaction and also carry the output of neurons and network inputs. The network function $f$ or output is given by a subset of node functions $\bar{y}$. The algorithm $a$ changes weights on edges.}
\label{fig:ANN}
\end{center}
\end{figure}

In an ANN instantiation of a CN, \emph{nodes} are neurons or units. Each neuron $i$ typically has a continuous state (output) determined by a function $y_i$ which is composed by two other functions: the weighted sum $S_i$ of its inputs $\bar{x}_i$ and an activation function $A_i$ such as the hyperbolic tangent. Directed \emph{edges} $ij$ (synapses) relate outputs $y_i$ of neurons $i$ to inputs $x_j$ of other neurons $j$, as well as external inputs and outputs with the network. Edges have a continuous state $w_{ij}$ (weight) that relates the states of neurons. The \emph{function} $f$ may be given by the states of a subset of $N$ (outputs $\bar{y}$), or by the complete set $N$. ANNs usually have two dynamical scales: a ``fast" scale where the network function $f$ is calculated by the functional composition of the function $y_i$ of each neuron $i$, and a ``slow" scale where an \emph{algorithm} $a$ adjusts the weights $w_{ij}$ (states) of edges. 
There is a broad diversity of algorithms $a$ used to update weights in different types of ANN. Figure \ref{fig:ANN} illustrates ANNs as CNs.

\section{Ant Colony Optimization}
\label{sec:3}

Ant colony optimization (ACO) is a population-based metaheuristic that can be used to find approximate solutions to difficult optimization problems \citep{Dorigo:2007}. ACO is inspired in the collective behavior of ants and their stigmergic interactions through pheromones.

\begin{figure}[htbp]
\begin{center}
\includegraphics[width=0.5\textwidth]{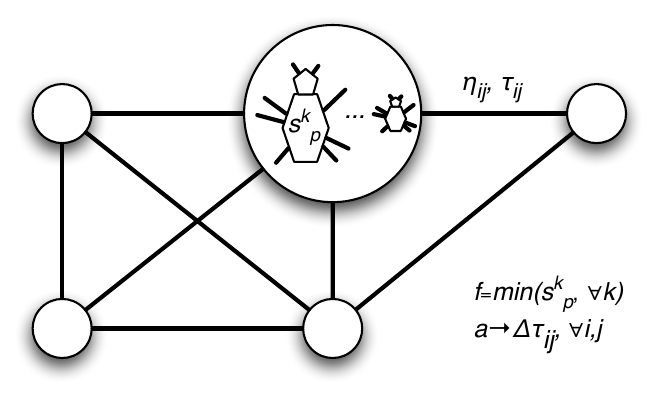}
\caption{Schematic of an ACO instantiation of a CN. Nodes contain ants that construct paths  $s_{k}^{p}$. Edges contain heuristic $\eta_{ij}$ and pheromone $\tau_{ij}$ values. The function $f$ is given by the best path found. Algorithm $a$ adjusts pheromone concentrations $\tau_{ij}$.}
\label{fig:ACO}
\end{center}
\end{figure}

In an ACO instantiation of a CN, \emph{nodes} are locations that contain a list of ``artificial ants" at their location. Each ant $k$ has  a path which represents a partial solution $s_{k}^{p}$, from which variables such as distance travelled and nodes visited can be extracted. \emph{Edges} (trails) have two variables: heuristic value $\eta_{ij}$ (e.g.~distance or cost between two nodes) and pheromone value $\tau_{ij}$. There have been different \emph{algorithms} proposed to calculate \emph{function} $f$, which is given by the shortest path found. As in ANNs, in ACO there are also two timescales: a ``fast" one in which ants travel through the network, generating paths (solutions) by choosing edges probabilistically at each visited node depending on their state $\eta_{ij}$,$\tau_{ij}$, and a ``slow" one, where the pheromone values $\tau_{ij}$ of edges are updated. This is similar to weight adjustment in ANNs. The pheromone update consists of an ``evaporation" phase, where all levels are reduced (similar to ``forgetting" in some ANNs) and an additive phase (similar to ``reinforcement" in some ANNs), where pheromone levels associated with good solutions are increased.
In some versions of ACO, there is a ``middle" scale, where ``demon" (problem specific) actions are taken, such as the application of a local search \citep{Dorigo:2007}.
Figure \ref{fig:ACO} illustrates ACO as a CN.

It can be argued that ACO---while inspired in the behavior of social insects~\citep{Garnier:2007}---does not serve as a realistic biological model. However, CNs can also be used to represent realistic models. Here the models of optimal decision-making presented by~\citet{Marshall:2009} are discussed. The problem of decision-making can be stated as choosing the best among two or more alternatives. It has been found that cortical neurons and social insects can approach an optimal balance between speed and accuracy in decision-making. Different individuals (neurons, insects, nodes in CNs) explore possibilities and interact (via synapses, pheromones, edges in CNs) to possibly change individual opinions. When a threshold is reached, i.e. enough individuals have made the same choice, the system selects that as a decision. The particularities (function $f$, algorithm $a$) of each model change, but all of them can be represented in terms of CNs. Here ACO is used as an example, but CNs can be used to compare more realistic models of swarms and brains.

\section{Particle Swarm Optimization}
\label{sec:4}

Particle swarm optimization (PSO) is a population-based stochastic approach for solving continuous and discrete optimization problems \citep{Dorigo:2008}. It was originally inspired by flocking algorithms \citep{reynolds87flocks} and social psychology research. In PSO, ``particles" move in a search space. Their position represents a candidate solution. Particles adjust their position and velocity depending on their neighboring particles in a graph.

\begin{figure}[htbp]
\begin{center}
\includegraphics[width=0.5\textwidth]{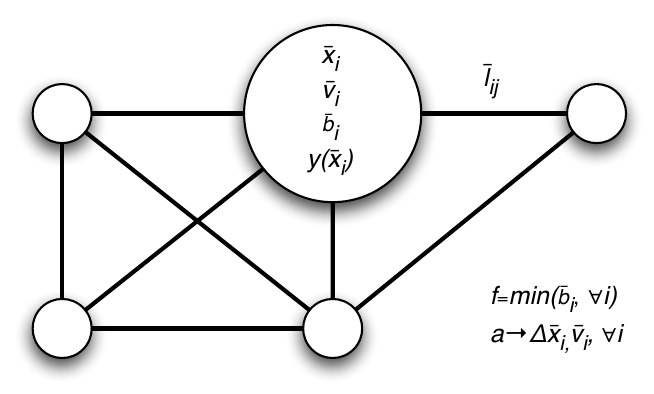}
\caption{Schematic of an PSO instantiation of a CN. Each node contains the  position $\bar{x}_i$ and velocity $\bar{v}_i$ of a particle, as well as its best solution found $\bar{b}_i$ and a function $y(\bar{x}_i)$. Edges contain the neighborhood's best solution $\bar{l}_{ij}$. Function $f$ is the best solution found by all particles. Algorithm $a$ changes the position $\bar{x}_i$ and velocity $\bar{v}_i$ of particles depending on the values of their neighbors.}
\label{fig:PSO}
\end{center}
\end{figure}

In a PSO instantiation of a CN, \emph{nodes} are particles with position $\bar{x}_i$, velocity $\bar{v}_i$, value of the best solution found $\bar{b}_i$, and a function $y(\bar{x}_i)$ that the network is trying to optimize. The position $\bar{x}_i$ represents a tentative solution. The function $f$ is simply the best solution found by $N$. \emph{Edges} represent the relationships between neighboring particles. Typically they contain information about the neighborhood's best solution, which can be represented as $\bar{l}_{ij}=max(\bar{b}_i,\bar{b}_j)$ for nodes $i,j$ related by edge $ij$. There is a variety of \emph{algorithms} to relate the way in which particles adjust their state. Again, two timescales can be identified: a ``fast" one, where particles evaluate the function they are trying to optimize ($y(\bar{x}_i)$), and a ``slow" one, where the velocity and position of particles are adjusted by algorithm $a$ depending on their previous states and those of their neighbors (links).
Figure \ref{fig:PSO} illustrates PSO as a CN.

For PSO, hypernetworks \citep{Johnson:2010} can be used as a generalization, so that a single edge can link more than two nodes and to represent the best solution of a neighborhood $\bar{l}_j$.

Like with ACO, PSO and other flocking algorithms are inspired in biology~\citep{reynolds87flocks,Rauch:1995}, but not quite realistic~\citep{Nagy:2010}. Nevertheless, CNs can also be used to model realistic models of of flocking. Here we focus on the model of flocking and schooling presented by~\citet{Couzin:2002}. Individuals (birds, fish, nodes) move in space, trying to maintain a minimum distance with their neighbors, i.e. avoid collisions. Also, individuals try to be attracted to their neighbors and align with them. A similar CN instantiation as the one shown in Figure \ref{fig:PSO} can be used to represent this model, which is considered to be realistic, even when it simplifies local interactions. More complex models~\citep{Nagy:2010}---where individuals have different interaction strengths according to their social heirarchy---can also be represented in terms of CNs. 

\section{Comparison and Discussion}
\label{sec:5}

Table \ref{table:comparison} shows a comparison of the language used to relate ANNs, ACO, and PSO in terms of CNs. It can be seen that all three architectures have the same basic components: nodes, edges, an algorithm, and a function. However, there are differences in the particularities of each architecture.

\begin{table}[htdp]
\caption{Particular instantiations of CNs: ANN, ACO, and PSO.}
\begin{center}
\begin{tabular}{|p{0.11\textwidth}|p{0.26\textwidth}|p{0.26\textwidth}|p{0.26\textwidth}|}
\hline
\textbf{CN} & \textbf{ANN} &	\textbf{ACO}	& \textbf{PSO}\\
\hline
Nodes	&	Neurons or units (function $y_i=A_i(S_i(\bar{x}_i))$)		&	Nodes (ants $k$ (path $s_{k}^{p}$))	&	Particles (position $\bar{x}_i$, velocity $\bar{v}_i$, best solution $\bar{b}_i$, function $y(\bar{x}_i)$) \\
\hline
Edges	&	Synapses (weight $w_{ij}$.)	&	Trails (heuristic value $\eta_{ij}$, pheromone concentration $\tau_{ij}$.) 	&	Relationships (neighborhood's best solution $\bar{l}_{ij}$)\\
\hline
Algorithm	&	Adjust edges ($\Delta w_{ij}$.)	&	Adjust edges ($\Delta \tau_{ij}$.)	&	Adjust nodes ($\Delta  \bar{x}_i$, $\Delta  \bar{v}_i$)\\
\hline
Function	&	Composition of functions of nodes	&	Shortest path ($min(s_{k}^{p})$)	&	Best solution ($min(\bar{b}_i)$)\\ 
\hline

\end{tabular}
\end{center}
\label{table:comparison}
\end{table}%

ACO and PSO have been used mainly for optimization. This explains why their $f$ is the minimum (best) of the solutions found. In contrast, ANNs have been used to solve many different tasks, e.g.~classification,  generalization, recognition, error correction, and time sequence retention. Still, all of the architectures can be described as computing a function $f$ in a distributed fashion. This is because they require the interaction of nodes to produce $f$.

It is interesting to note that, even when ACO and PSO are inspired by swarming systems, algorithms of ANN and ACO are more similar between themselves than with PSO, in the sense that they update edges, while PSO algorithms update nodes. However, the models can be extended from networks to hypernetworks \citep{Johnson:2010}, where there is a duality between nodes and edges, i.e.~one can exchange nodes and edges while preserving the functionality of the hypernetwork. In this case, PSO particles can be described as hyperedges, and their interactions as nodes. Then, the PSO algorithm $a$ would update hyperedges.



\subsection{Dynamical scales}

One common characteristic among all three architectures studied is that they have ``slow" ($a$) and ``fast" ($f$) dynamical scales. This is no coincidence. Having multiple dynamical scales is a requirement for computing complex functions\footnote{A complex function will not be defined formally, but it can be understood as a  function that is non-trivially described, explored or optimized.} that change in time, i.e. are non-stationary. If there is only change at a single scale, then the phase space of $f$, i.e. all its possible values and potentially its optimum, can be explored, but it cannot be changed. Having two dynamical scales, the changes in the phase space of $f$ can be explored as well. This property is essential when $f$ is not known beforehand: the algorithm explores different phase spaces until one that satisfies $f$ is found. 

The tasks solved by real neural and swarm systems also need to exploit the advantages of multiple dynamical scales. In the case of neural systems, learning (synapse modification) enables the correct adjustment of a particular function of a circuit, e.g.~categorization. For swarming insects, local interactions (direct or stigmergic) enable the colony to make complex decisions, e.g.~choosing a new nest.

\begin{figure}[htbp]
\begin{center}
\includegraphics[width=0.9\textwidth]{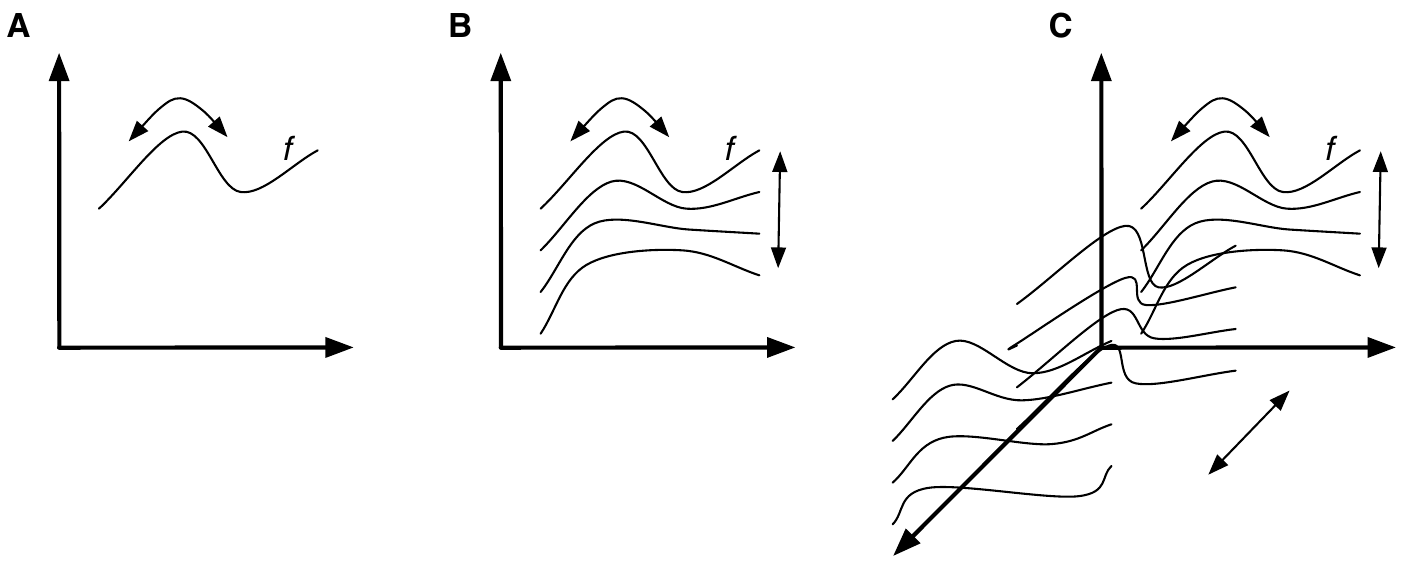}
\caption{Changes at different dynamical scales: (A) single scale: values can vary only along $f$, (B) double scale: apart from changes along $f$, $f$ can also be varied, and (C) triple scale: changes in ways in which $f$ can be varied can also be explored. Note that these diagrams are only illustrative. $f$ can certainly be multidimensional, i.e.~in $\mathbb{R}^n$.}
\label{fig:scales}
\end{center}
\end{figure}

Would it be useful to have three dynamical scales? This would imply the exploration of changes in the space of phase spaces of $f$. For example, this is used in ``evo-devo" \citep{fontana2002modelling,Munteanu:2008} or epigenetic \citep{EpiRob2001} algorithms, where there is a function $f$, its phase space is explored through the ``lifetime" of an ``organism" (learning), and the space of possible organisms is explored at an evolutionary scale, e.g.~with evolutionary algorithms. An example can be seen with the work of \citet{Botee:1998}, where a genetic algorithm is used to find the best parameters of ACO.
Figure \ref{fig:scales} illustrates the change possible at one, two and three dynamical scales.

It should be noted that multiple dynamical scales are an important feature to enable \emph{adaptation} \citep{Holland1975,Holland1995}. A system can function at a ``fast" scale, while adaptation can work at a ``slow" scale. When the situation of the system changes, adaptation can change the function of the system to cope with the new situation.

A question that arises is whether CNs with three dynamical scales are computationally equivalent, or more powerful, than CNs with two dynamical scales. The reader is invited to ponder on this question, which is already out of the scope of this paper.

\subsection{Functional scales and the relevance of interactions}

Apart from having multiple dynamical scales, CNs have multiple functional scales. The most clear scales are those of node (local) and network (global). Subnetworks, modules, layers, or motifs can also form intermediate scales. In CNs, nodes compute certain ``local" functions. These functions are combined to produce the CN's ``global" function $f$. However, $f$ cannot be \emph{reduced} to the node functions alone. Since the states of the nodes depend on other nodes, \emph{interactions} are relevant to determine the future state of nodes, and thus $f$.

As in the case of dynamical scales, having multiple functional scales is a requirement for computing complex functions. In this context, interactions can be described as operators. Local structures (e.g.~nodes, motifs) can store certain information and can compute certain functions. However, in many cases, the information produced by local structures is less complex than the one that produced by the global structure (i.e.~network). This is because the interactions between local structures integrate information produced at the lower scales to compute the global $f$. The exceptions are trivial, e.g.~when all the interactions are weak or absent, or the local structures are redundant. In these cases, one can say that the complexity of the local structures is the same as the complexity of the global one.

This will be clearer introducing a definition of what is meant by complexity: \emph{Complexity is the amount of information necessary to describe a phenomenon at a particular scale} \citep{BarYam2004,Prokopenko:2008,Gershenson:2007}. With a CN, in most cases more information is necessary to describe the whole network than the collection of all its nodes, namely because of the information contained in edges, which represent interactions. Repeating what was stated above, $f$ cannot be reduced to $N$ only, namely because of $K$.

\begin{figure}[htbp]
\begin{center}
\includegraphics[width=0.9\textwidth]{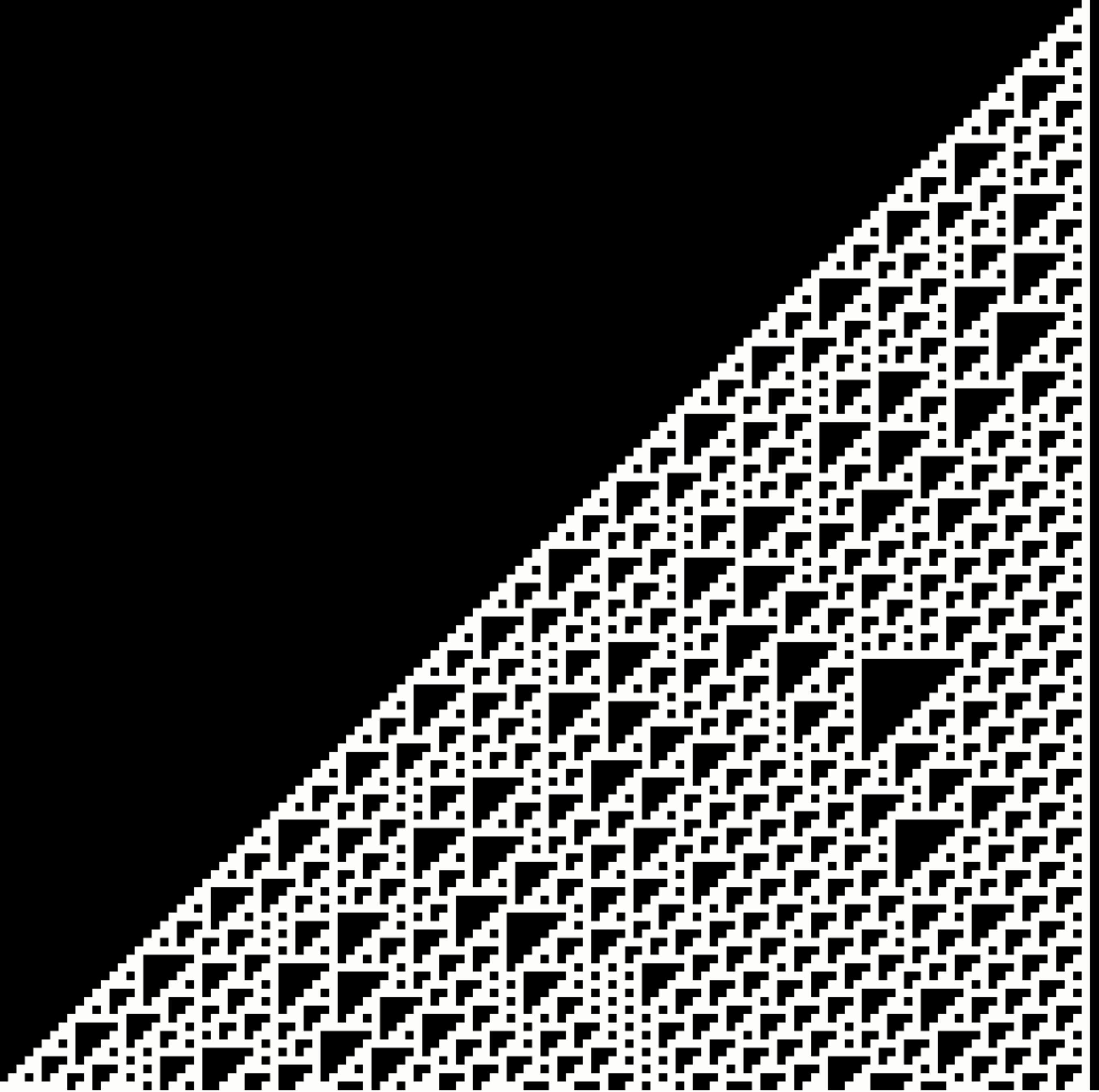}
\caption{Temporal evolution of ECA 110. Each cell is represented by a column and time flows downwards, i.e.~each row represents the state of the CA at successive time steps. Black cells represent '0' and white cells represent '1'. The first row (initial state) consists of a single '1'. The state of other rows depends on the state of the row above. It is not possible to compute \emph{a priori} the state of the last row from the first row without computing all the intermediate states.}
\label{fig:ECA110}
\end{center}
\end{figure}

A clear example of the relevance of interactions can be seen with cellular automata (CA) \citep{vonNeumann1966,Wolfram1986,WuenscheLesser1992,Wolfram:2002}, which incidentally can also be described in terms of CNs. The states of cells (nodes) depend on the state of their neighbors (edges) according to a certain rule. In the case of elementary cellular automata (ECA) 110 \citep{Wolfram:2002,Juarez:2007}, the state at time $t+1$ of each cell depends on its state and of its closest neighbors (3 cells in total) at time $t$. The updating is done synchronously according to the values shown in Table \ref{table:ECA110}. Figure \ref{fig:ECA110} shows the temporal evolution of ECA 110 for a particular initial state. Even when the behavior of ECA 110 is determined by very simple rules, it is capable of universal computation \citep{Cook2004}, exploiting the interactions between emergent structures \citep{Juarez:2007} (slow scale) that arise from the simple interactions (fast scale) of the local neighborhoods.

\begin{table}[htdp]
\caption{ECA 110 lookup table. The first column shows the eight possible states of the 3 cells used to update every cell, while the second column shows the state of the updated cell.}
\label{table:ECA110}
\begin{center}
\begin{tabular}{|c|c|}
\hline
$t$	&$t+1$\\
\hline
000	&0	\\
\hline
001	&1	\\
\hline
010	&1	\\
\hline
011	&1	\\
\hline
100	&0	\\
\hline
101	&1	\\
\hline
110	&1	\\
\hline
111	&0	\\
\hline
\end{tabular}
\end{center}
\end{table}%

With ECA 110, the relevance of interactions is clearly seen. CNs with simple nodes and functions are capable of complex computations because of the relevant information contained in edges. Note that interactions are not necessarily physical, but they are real. For different systems, there are different ``implementations" of edges, e.g.~synapses, pheromones, or cues \citep{Couzin:2009}. Still, they all have the same role: to relate states of nodes to compute a distributed function $f$. Using the CN formalism, it can be explained how the computational power of a brain is much more complex than that of a large collection of isolated neurons, and the computational power of a swarm is much more complex than that of a group of isolated insects. Not only interactions are important, but also multiple dynamical and functional scales.

For functional scales, we can also ask whether only two scales are less powerful than more than two scales. However, again, the question is beyond the scope of this paper, although many have discussed the advantages of \emph{modularity}~\citep{Simon1996,Schlosser:2004,Callebaut:2005}.

\subsection{Which architecture is the best?}

One might wonder which architecture---ANNs, ACO, or PSO---is the best. There is no best architecture independently of a specific context \citep{wolpert95no,wolpert1997no,Gershenson2004}. Different implementations of CNs will be more adequate for different problems, either giving better solutions, or improved speed. The convenience of a particular architecture does not depend only on the problem: different methods will be more useful for different people, depending on their experience and expertise.

A valid question would be: which architecture---ANNs, ACO, or PSO---is more computationally powerful? Since the architectures are so general, it can be conjectured that they are computationally equivalent. For example, one could implement e.g.~an ANN based on ACO or PSO, e.g.~where the function of a node is itself determined by an ACO or PSO CN. Similarly, one can implement an ACO or PSO based on ANNs. Finally, one can also develop ACO based on PSO and vice versa. It might not be useful at all, but the idea shows that computationally (in Turing's \citeyearpar{Turing:1936} sense) they all have similar capacities. A formal proof of this conjecture is beyond the scope of this paper. There will be more differences on particular implementations of ANNs (e.g.~given by number of nodes and edges) than between a given ANN and an equivalent ACO or PSO.

The literature is rich in examples of hybrid systems, where some properties of one architecture are combined with those of another one, e.g.~\citep{Kennedy:1995,Wang04particleswarm,Chen:2004,Blum:2005,Mozafari:2006,Martin:2010} to cite a few of them. Actually, the original PSO paper \citep{Kennedy:1995} used PSO as an example to train an ANN.
This illustrates that for a particular problem and for a particular expertise of the developers, no single approach gives the best solutions.

Having discussed the similarity of the computational capacities of neural and swarm architectures, we can continue with the discussion about the role of the architectures in cognition.

\subsection{Cognition}

Cognition comes from the Latin \emph{cognoscere}, which means ``get to know". We can say that \textbf{a system is cognitive if it knows something} \citep{Gershenson2004}. With this definition, it is not possible to draw a boundary between cognitive and non-cognitive systems. Since somebody has to judge whether a system knows or not, it is partly observer-dependent. Instead of discussing whether a system is cognitive or not, it is more fruitful to distinguish different types of cognition (e.g.~human, animal, biological (including plant and bacterial), social, artificial, adaptive, systemic \citep{Gershenson2004}), to compare and better understand them.

From this perspective, it is clear that swarms are cognitive systems because they \emph{know} how to forage, find sites, build nests, and even add and subtract small numbers \citep{Reznikova:2007,Ryabko:2009}. Neural architectures are cognitive because they \emph{know} how to categorize, classify, remember, etc. \citep{hopfield1982neural}. To compare both types of cognition, we can use the concept of computing networks proposed in this paper.

\textbf{Cognition can be seen as the ability to compute a function $f$}. This is because if a system can compute $f$, we can say that it \emph{knows} how to calculate $f$. This vocabulary does not aim at ascribing to CNs a ``mind", ``consciousness", or other difficult-to-define property usually associated with human cognition. The aim of this use of language is to be able to compare the cognitive capacities of neural and swarm architectures. As discussed in the previous subsection, neural and swarm architectures have similar computational abilities, shown by their generalization as CNs. If we describe cognition as computation, it naturally follows that neural and swarm architectures have similar cognitive capacities, \emph{in theory}. In practice, different implementations will have different cognitive abilities, just as a human brain has different abilities as a rat brain: the former is potentially better at poetry, the latter is potentially better at navigation. Also, differences of timescale are important, i.e. brains usually compute at faster timescales than swarms.

The great advantage of swarm and neural cognition is that they manage to exploit the benefits of multiple functional and dynamical scales to exhibit complex cognitive abilities. As discussed above, multiple scales enable CNs to compute more complex functions and to adapt to changes in the environment. In cognitive terms, the structure represented by CNs enables neural and swarm architectures to exhibit a more complex cognition, as compared to a system with a single functional or dynamical scale, e.g.~a thermostat. We can see that there are cognitive systems with more than two scales, e.g.~group cognition \citep{Stahl2006}, which exploit and combine the cognitive abilities of a collection of humans. Naturally, swarms are another example of multiple scale cognition, since the cognition of individual insects is provided by a neural architecture.

\subsection{Alternative descriptions}

The description of ANN, ACO, and PSO in terms of computing networks is only one of several possible languages that can be used to compare the architectures. For example, a \emph{multi-agent} description can be also used: Nodes can be described as agents and edges can be described as interactions. An algorithm regulates the interactions between agents to reach a global state (equivalent to function $f$). This global state can be described as being reached by self-organization \citep{GershensonHeylighen2003a}. This self-organization in a multi-agent system is comparable to the distributed computation of $f$. The system can compute the same function $f$, only the description changes. For the purposes pursued in this paper, the network description seems more appropriate. A multi-agent description can be valuable in the process of designing algorithms, since goals of agents and systems can be defined. Then, the algorithm should minimize ``friction" (i.e.~negative interactions) and promote ``synergy" (positive interactions) \citep{GershensonDCSOS}. This will necessarily increase the system's ``satisfaction", which is basically what we want the system to do, i.e.~$f$.

Yet another description that can be used is that of \emph{information} \citep{Gershenson:2007}. Nodes, edges, algorithms, and functions can be all seen as information, while computation can be seen as a change of information. This is a more general description, so it is not so useful for making a comparison as the one presented here. The information framework might be useful for finding general principles across disciplines, since everything can be described in terms of information.

Neural and swarm architectures can also be described in terms of differential equations, dynamical systems theory, object-oriented programming, rules, zeros and ones, etc.
Different descriptions are suitable for different contexts and purposes \citep{Gershenson2004}. The purpose of computing networks is specifically the comparison of neural and swarm architectures. 
CNs will not be as good as the original descriptions for developing e.g.~new learning algorithms in ANNs or new optimization algorithms in ACO. This is because the computing networks description is more general and vague than an actual instantiation of an ANN or PSO. More details are required at the implementation level, which were neglected here. The goal of defining CNs is more theoretical than practical: to understand the similarities and differences of neural and swarm architectures, not to improve current technical algorithms. CNs are not better or worse that other descriptions. Here they were useful to understand the relevance of multiple scales and some computational principles common to neural and swarm architectures at a general level. It might have been made with a different description, but CNs seemed the most appropriate for the purposes of this work.

\section{Conclusions}
\label{sec:6}

As Trianni and Tuci suggest \citep{Trianni-Tuci:09:ecal}, the principles of swarms can be useful tools for studying the neuroscientific basis of cognition. Here it was shown that both swarm and neural architectures share similar computational and cognitive abilities. This was achieved by defining computing networks (CNs), which are able to generalize neural and swarm architectures, allowing their comparison. CNs can also be useful to generalize and compare other swarm intelligence algorithms, e.g. \citep{pham2006bees,Yang09,Krishnanand:2009}.
By studying the general principles that enable CNs to perform complex computations, one can understand better what are the requirements of neural and swarm systems to exhibit complex cognition. In this paper, the importance of having multiple dynamical and functional scales to exhibit complex cognition and adaptation was discussed. From a cognitive perspective, CNs support the thesis of neural and swarm architectures having similar cognitive abilities. CNs also show that neural and swarm architecture have similar computational abilities.

\section*{Acknowledgements}
I should like to thank Dante Chialvo, Vito Trianni, Elio Tuci, Tam\'as Vicsek, and anonymous referees for useful comments. This work was partially supported by SNI membership 47907 of CONACyT, Mexico.

\bibliographystyle{cgg}
\bibliography{carlos,swarmCognition,RBN,sos,COG,complex}

\end{document}